\pdfoutput=1

\documentclass[11pt]{article}

\usepackage{acl}

\usepackage{times}
\usepackage{amsmath}
\usepackage{amssymb}
\usepackage{latexsym}
\usepackage{graphicx}
\usepackage[T1]{fontenc}
\usepackage{booktabs} 
\usepackage{multirow}  

\usepackage[utf8]{inputenc}

\usepackage{microtype}

\usepackage{lipsum}
\usepackage{tcolorbox}
\definecolor{pastelgreen}{RGB}{221, 245, 193}
\definecolor{pastelred}{RGB}{245, 201, 197}

\tcbset{
   pluscolorbox/.style={
      colframe=pastelred,
      colback=pastelred,
      boxrule=0.5pt,
      arc=2pt,
      boxsep=0pt,
      left=2pt,
      right=2pt,
      top=2pt,
      bottom=2pt
   },
   minuscolorbox/.style={
      colframe=pastelgreen,
      colback=pastelgreen,
      boxrule=0.5pt,
      arc=2pt,
      boxsep=0pt,
      left=2pt,
      right=2pt,
      top=2pt,
      bottom=2pt,
   }
}

\newtcbox{\redcolorbox}{pluscolorbox}
\newtcbox{\greencolorbox}{minuscolorbox}

\newcommand{\plusce}[2]{$#1_{\mbox{\tiny{\redcolorbox{↑#2}}}}$}
\newcommand{\minusce}[2]{$#1_{\mbox{\tiny{\greencolorbox{↓#2}}}}$}

\newcommand{\plus}[2]{$#1_{\mbox{\tiny{\greencolorbox{↑#2}}}}$}
\newcommand{\minus}[2]{$#1_{\mbox{\tiny{\redcolorbox{↓#2}}}}$}
\newcommand{\greydots}{\color{gray}$\cdots\cdots\cdots\cdots\cdots\cdots\cdots\cdots\cdots\cdots\cdots\cdots\cdots\cdots\cdots\cdots\cdots\cdots\cdots\cdots\cdots\cdots\cdots\cdots\cdots\cdots\cdots\cdots\cdots\cdots\cdots\cdots\cdots\cdots\cdots\cdots\cdots\cdots\cdots\cdots\cdots\cdots\cdots\cdots\cdots\cdots\cdots\cdots\cdots\cdots$
}
\title{When Quantization Affects Confidence of Large Language Models?}

\author{
    ~\textbf{Irina Proskurina}, 
    ~\textbf{Luc Brun},
    ~\textbf{Guillaume Metzler},
    ~\textbf{Julien Velcin}
 \\
   Universit{\'e} de Lyon, Lyon 2, ERIC UR 3083, France\\     \textbf{Correspondence:} \href{mailto:Irina.Proskurina@univ-lyon2.fr}{Irina.Proskurina@univ-lyon2.fr}
}

\begin{document}
\maketitle
\begin{abstract}
Recent studies introduced effective compression techniques for Large Language Models (LLMs) via post-training quantization or low-bit weight representation. 
Although quantized weights offer storage efficiency and allow for faster inference, existing works have indicated that quantization might compromise performance and exacerbate biases in LLMs.
This study investigates the confidence and calibration of quantized models, considering factors such as language model type and scale as contributors to quantization loss.
Firstly, we reveal that quantization with GPTQ to 4-bit results in a decrease in confidence regarding true labels, with varying impacts observed among different language models. 
Secondly, we observe fluctuations in the impact on confidence across different scales. 
Finally, we propose an explanation for quantization loss based on confidence levels, indicating that quantization disproportionately affects samples where the full model exhibited low confidence levels in the first place.
We make our code and quantized models publicly available.\footnote{\url{https://github.com/upunaprosk/quantized-lm-confidence}}
\end{abstract}

\section{Introduction}

Large language models (LLMs) are widely used in a variety of natural language generation applications \cite{Bahdanau2014NeuralMT, brown2020language, winata-etal-2021-language, scao2022bloom, touvron2023llama}.
LLMs have been proven to achieve high performance in zero and few-shot prompting, providing results on par with fine-tuned baselines, especially in commonsense reasoning tasks~\cite{zhang2022opt,scao2022bloom,jiang2023mistral}.
\citealp{kaplan2020scaling} show that emerging abilities come with the scale increase, which makes well-performing larger models less accessible and limits their practical usability.
A range of efficient compression and acceleration methods, including quantization, have been developed that help to alleviate high latency and extensive storage demands~\cite{Gupta2020CompressionOD,tao-etal-2022-compression}. 
Despite its efficacy as a compression technique, recent works show that quantization may degrade the initial performance and amplify the sensitivity of an LLM to certain linguistic phenomena and stereotypes~\cite{liu2023emergent,ramesh-etal-2023-comparative}. 
However, less attention has been paid to explaining the compression loss, particularly its variance across different texts.
\begin{figure}
  \centering
  \includegraphics[width=0.95\linewidth]{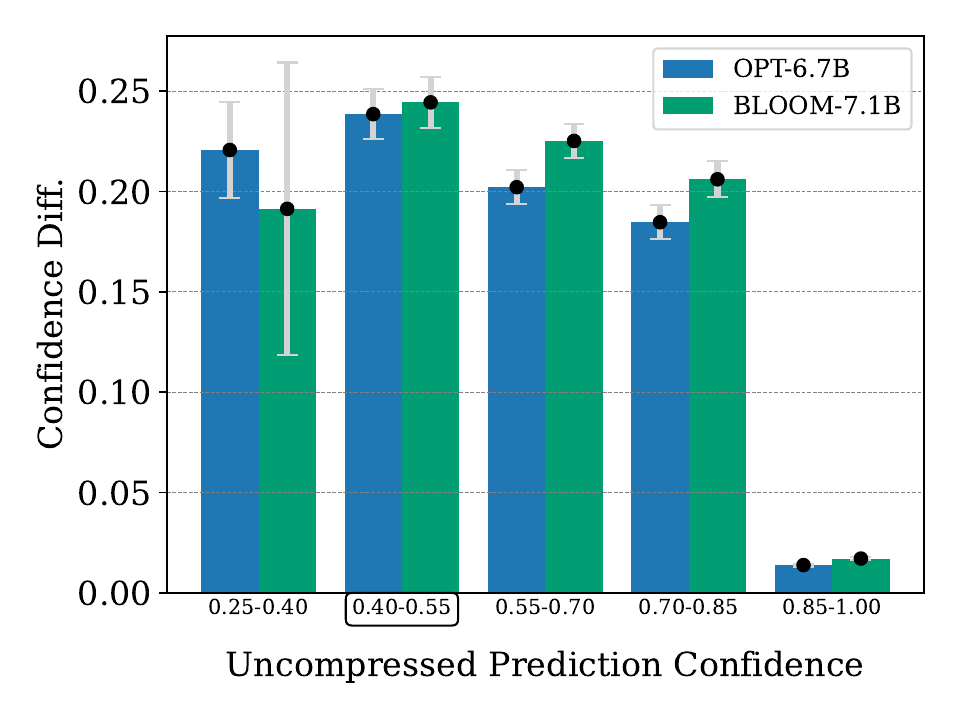}
  \caption{ 
    Quantization-induced absolute confidence shifts in original (pre-compression) low and high confidence samples (BLOOM and OPT models, \textsc{HellaSwag} benchmark). The bin with the largest mean confidence shift is highlighted.
  }
  \label{fig:conf_diffs}
\end{figure}
In this paper, we extend the existing research on the compression loss estimation; in particular, we measure the impact of quantization on the confidence of LLMs that can be initially overconfident in both right and wrong predictions~\cite{jiang-etal-2021-know,xiao-etal-2022-uncertainty,ahuja-etal-2022-calibration,desai-durrett-2020-calibration}.


\begin{table*}[th!]
{
    \centering
    \footnotesize
    \resizebox{0.96\textwidth}{!}{
        \begin{tabular}{l|cc|cc|cc|cc|cc|cc} 
            \toprule
            \multirow{2}{*}{\textbf{Model}} &
            \multicolumn{2}{c|}{\rotatebox{0}{\textbf{\textsc{Arc Easy}}}} &
            \multicolumn{2}{c|}{\rotatebox{0}{\textbf{\textsc{BoolQ}}}} &
            \multicolumn{2}{c|}{\rotatebox{0}{\textbf{\textsc{HellaSwag}}}} &
            \multicolumn{2}{c|}{\rotatebox{0}{\textbf{\textsc{OpenBookQA}}}} &
            \multicolumn{2}{c|}{\rotatebox{0}{\textbf{\textsc{PiQA}}}} &
            \multicolumn{2}{c}{\rotatebox{0}{\textbf{\textsc{XStory}}}} \\
            \cmidrule{2-13}
            & \textbf{Acc.$\uparrow$} & \textbf{CE$\downarrow$} & \textbf{Acc.$\uparrow$} & \textbf{CE$\downarrow$} & \textbf{Acc.$\uparrow$} & \textbf{CE$\downarrow$} & \textbf{Acc.$\uparrow$} & \textbf{CE$\downarrow$} & \textbf{Acc.$\uparrow$} & \textbf{CE$\downarrow$} & \textbf{Acc.$\uparrow$} & \textbf{CE$\downarrow$} \\
            \midrule
            {\includegraphics[height=0.4cm]{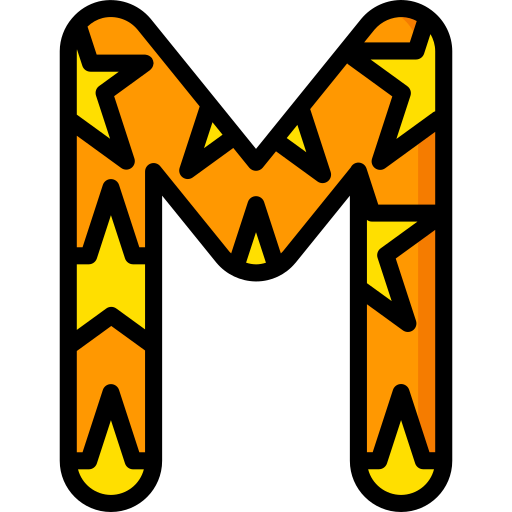}}{\color{gray}$\cdots$} & \multicolumn{12}{c}{\greydots} \\
            7B & \minus{ \textbf{81.10} }{ 1.18 } & \plusce{\textbf{7.94}}{0.83} & \minus{\textbf{ 83.61} }{ 0.86 } & \plusce{38.62}{3.13} & \minus{\textbf{ 61.30} }{ 1.53 } & \plusce{\textbf{34.3}}{1.29} & \minus{ \underline{32.60} }{ 0.40 } & \plusce{\underline{45.24}}{2.08} & \minus{ \textbf{80.83} }{ 0.65 } & \minusce{45.24}{0.4} & \minus{ \textbf{78.89} }{ 0.27 } & \minusce{\textbf{4.78}}{0.08} \\
            {\includegraphics[height=0.5cm]{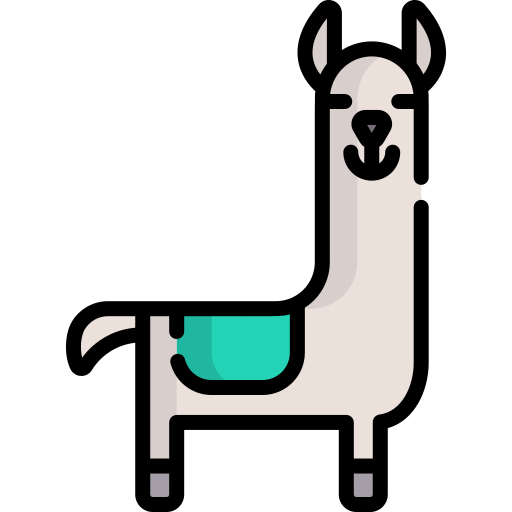}}{\color{gray}$\cdots$} & \multicolumn{12}{c}{\greydots}\\
            7B & \minus{\underline{75.25}}{4.29} & \plusce{\underline{9.99}}{1.72} & \minus{\underline{75.05}}{2.51} & \minusce{38.78}{7.66} & \minus{\underline{56.94}}{3.23} & \plusce{\underline{37.8}}{4.07} & \minus{\textbf{34.0}}{4.2} & \plusce{\textbf{44.56}}{3.13} & \minus{\underline{78.67}}{2.01} & \minusce{44.94}{0.58} & \minus{\underline{76.77}}{2.05} & \plusce{\underline{4.97}}{0.24}\\
            {\includegraphics[height=0.5cm]{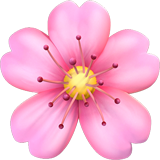}}{\color{gray}$\cdots$} & \multicolumn{12}{c}{\greydots} \\
            560M & \minus{ 47.35}{1.56} & \plusce{29.13}{1.13} & \minus{55.14}{6.94}& \minusce{26.91}{2.85} & \minus{31.58}{0.47}& \minusce{64.81}{0.49}& \minus{17.2}{0.6} &\plusce{61.16}{ 1.61}&\minus{64.09}{1.14}&\plusce{\textbf{40.98}}{0.32}&\minus{61.22}{1.39}&\minusce{5.13}{0.03} \\
            1.1B & \minus{51.47}{2.27} & \plusce{25.07}{2.17} & \plus{59.08}{0.74} &\plusce{32.8}{2.65} &\minus{34.44}{0.87} & \plusce{58.51}{0.85} & \minus{20.0}{2.2} & \plusce{58.88}{0.74} & \minus{67.14}{0.98} & \minusce{42.27}{0.68} & \minus{62.54}{1.52}& \plusce{5.77}{0.15} \\
            1.7B & \minus{56.31}{1.81} & \plusce{21.99}{0.35} & \minus{61.77}{0.12} & \minusce{38.29}{0.06} & \minus{37.54}{0.82} & \plusce{55.67}{0.51} & \minus{ 21.40 }{ 1.60 } & \plusce{56.64}{0.97} & \minus{68.77}{0.76} & \minusce{\underline{41.4}}{0.49} & \minus{64.66}{0.53} & \plusce{5.65}{0.08}\\
            3B & \minus{5947}{2.27} & \plusce{19.68}{1.31} & \plus{61.62}{0.09} & \minusce{34.67}{0.86} & \minus{41.39}{0.91} & \plusce{52.33}{0.82} & \minus{ 21.6 }{ 0.40 } & \minusce{56.32}{0.15} & \minus{70.84}{0.82} & \minusce{42.12}{0.25} & \minus{66.78}{0.53} & \plusce{5.76}{0.2} \\
            7.1B & \minus{65.03}{1.56} & \plusce{15.57}{1.06} & \plus{62.81}{0.19} & \plusce{32.28}{0.19} & \minus{46.49}{1.11}& \plusce{48.54}{1.07} & \minus{ 25.20 }{ 0.80 } & \plusce{53.23}{0.01} & \plus{72.63}{0.28} & \minusce{42.52}{0.24} & \minus{70.55}{0.33} & \minusce{5.53}{0.1}\\
            {\includegraphics[height=0.5cm]{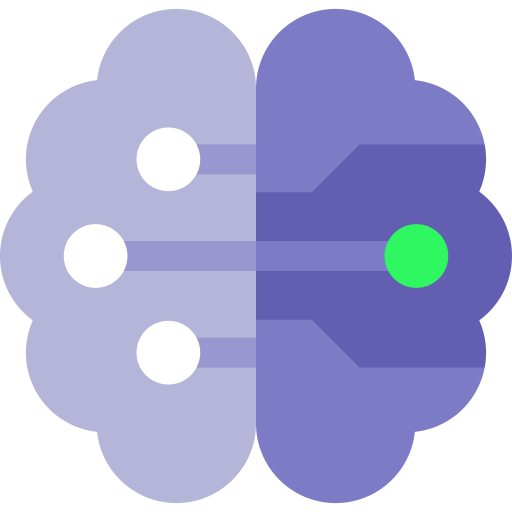}}{\color{gray}$\cdots$} & \multicolumn{12}{c}{\greydots} \\ 125M & \minus{ 43.56 }{ 1.05 } & \plusce{32.76}{0.62} & \minus{ 55.44 }{ 2.72 } & \minusce{30.13}{3.48} & \minus{ 29.18 }{ 0.59 } & \plusce{62.84}{0.48} & \plus{ 16.6 }{ 0.20 }& \plusce{61.19}{0.45} & \minus{ 62.00 }{ 0.81 } & \plusce{41.51}{0.11} & \minus{ 58.84 }{ 1.0 } & \minusce{5.9}{0.12}\\
            350M & \plus{ 44.20 }{ 0.08 } & \plusce{31.21}{1.79} & \minus{ 57.65 }{ 3.46 } & \minusce{29.62}{4.4} & \minus{ 32.02 }{ 0.18 } & \plusce{60.09}{0.30} & \minus{ 17.60 }{ 1.40 } & \minusce{61.92}{0.19}  & \minus{ 64.47 }{ 0.92 } & \plusce{41.58}{0.19} &\minus{ 62.48 }{ 1.79 }& \minusce{5.97}{0.35}\\
            1.3B & \minus{ 56.99 }{ 0.85 }& \plusce{20.85}{0.85} &\minus{ 57.67 }{ 2.73 } & \plusce{\underline{26.42}}{0.01} & \minus{ 41.56 }{1.00} & \plusce{52.7}{0.16} & \minus{ 23.4 }{ 1.40 } & \plusce{55.04}{3.32} & \minus{ 71.71 }{ 0.76 } & \plusce{41.49}{0.34} & \minus{ 70.28 }{ 1.58 }& \plusce{5.6}{0.23}\\
            2.7B & \minus{ 60.77 }{ 1.64 } & \plusce{17.63}{1.56} & \minus{ 60.24 }{ 5.05 } & \plusce{\textbf{25.86}}{0.01} & \minus{ 45.86 }{ 0.51 } & \minusce{48.93}{0.15} & \minus{ 25.0 }{ 2.20 } & \plusce{52.6}{2.76} & \minus{ 73.78 }{ 0.98 } & \plusce{41.87}{0.39} & \plus{ 70.42 }{ 0.13 } & \minusce{5.83}{0.06}\\
            6.7B & \minus{ 65.57 }{ 0.08 }& \minusce{15.58}{0.18} & \minus{ 66.05 }{ 0.83 } & \minusce{28.05}{1.72} & \minus{ 50.51 }{ 0.65 } & \plusce{45.25}{0.01} & \minus{ 27.6 }{ 1.20 } & \plusce{50.99}{1.44}  & \minus{ 76.28 }{ 0.22 }& \minusce{43.72}{0.65} & \minus{ 73.6 }{ 0.19 } & \minusce{5.62}{0.17} \\
            13B & \plus{ 67.13 }{ 0.38 } & \minusce{14.21}{0.69} & \minus{ 65.93 }{ 0.09 } & \minusce{29.47}{0.52} & \minus{ 52.43 }{ 0.59 } & \plusce{43.03}{0.47} &  \minus{ 27.2 }{ 0.05 } & \minusce{52.33}{0.23}  & \plus{ 75.84 }{ 0.11 } & \minusce{43.87}{0.43} & \minus{ 76.04 }{ 0.07 } & \plusce{5.15}{0.21} \\
            \bottomrule
        \end{tabular}}}
    \caption{\label{tab:calibration_score} 
    Zero-shot accuracy scores (Acc.) and calibration error (CE) for full LLMs by benchmark with the difference in scores after quantization. We report expected CE for binary tasks and adaptive CE for multi-class benchmarks (\textsc{ARC}, \textsc{BoolQ}, \textsc{OpenBookQA}). \underline{Notations:}
    \includegraphics[height=0.35cm]{images/mistral.png}=\textsc{Mistral}; \includegraphics[height=0.35cm]{images/llama.png}=\textsc{LLaMA};\includegraphics[height=0.35cm]{images/cherry_blossom.png}=\textsc{BLOOM};\includegraphics[height=0.35cm]{images/opt3.png}=\textsc{OPT}.}
\end{table*}

\textbf{Our main contributions} are the following: (i)~we investigate how quantization with GPTQ~\cite{frantar2022gptq} influences the calibration and confidence of LLMs, (ii)~we assess the confidence alignment between compressed and full LLMs at scale, (iii)~we explain the quantization loss from the initial confidence perspective.

Our null hypothesis is that the compressed vs. full predictive probability distributions are indistinguishable since prior work discussed a negligible accuracy drop in performance after quantization~\cite{jacob2018quantization,dettmers2022llm,xiao2023smoothquant}.
We analyze the relationship between models by comparing calibration scores—indicating a model's ability to accurately reflect true probabilities—before and after quantization.
To the best of our knowledge, our research is the
first attempt to explain the quantization loss through the lens of predictive probabilities.

\section{Related Work}
The pretrained knowledge embedded in very large models has paved the way to parameter-efficient adaptation for downstream tasks, such as prompting and few-shot learning, bypassing the necessity for fine-tuning~\cite{brown2020language,wei2022emergent}.
The inference of LLMs can be accelerated through a low-bit representation of trained weights (\textit{quantization}) and effective tensor slicing across multiple GPUs (\textsc{DeepSpeed}~\cite{rasley2020deepspeed}, \textsc{Accelerate}~\cite{accelerate}, \textit{inter alia}).
Prior studies have estimated compression efficiency through:
(1) latency-related measures determining throughput and a multiple of the original model's inference speed-up, (2) the precision of weights approximation, and (3) performance decrease (gap)~\cite{jacob2018quantization,dettmers2022llm,xiao2023smoothquant,frantar2022gptq}.
Recent comparative studies on interpreting compression loss have indicated that compression amplifies biases and stereotypes, highlighting a disparate quantization loss in multilingual LLMs across different architectures~\cite{ramesh-etal-2023-comparative}.
In contrast, another line of research suggests that compression enhances fairness~\cite{hessenthaler-etal-2022-bridging}.
Altogether, existing studies commonly measure compression loss by observing the deviation in performance before and after quantization.
In this project, we adopt the recent GPTQ quantization method for compressing model weights and concentrate on the disparities between predictive probability distributions instead.
For the first time, our approach reveals the relationship between the initial level of predictive confidence and quantization loss. 

\section{Methodology}
We follow \citealp{jiang2021can} and consider a classification problem where inputs to the model are questions $x$ paired with candidate answers $y$ to constitute concatenated sequences.
The generative model then processes these concatenated question-answer pairs to predict the most probable answer $\hat{y}$ from the provided choices $Y$ for a given $x$:
\begin{align*} 
\hat{y} = \underset{y \in Y}{\text{arg max }} p_{\text{LM}}(y|x).
\end{align*}
Here, the probability of the token sequence 
$y$ is derived as the product of individual token $y_{[i]}$ probabilities within the sequence, conditioned on 
$x$ and the preceding tokens $y_{[1:i-1]}$:
\begin{align*} 
p_{\text{LM}}(y|x) = \prod_{i=1}^{|y|} p_{\text{LM}}(y_{[i]}|x, y_{[1:i-1]}),
\end{align*}
where $|y|$ is the number of tokens composing the answer $y$.

For the entailment generation benchmarks, we use texts concatenated with possible completions as inputs to the model. 
We compare the quantized and full-precision models with the difference in the probabilities of the sequences  $p_{\text{LM}}(y|x)$, further referred to as confidences.
\subsection{Quantization}

We quantize pre-trained weights of LLMs with a post-training quantization method known as GPTQ (OPTQ, \citealp{frantar2022optq}).
This approach employs iterative layer-wise weight quantization based on the input data, providing several benefits compared to other quantization methods: minimized weight approximation error, support for serialization across various bit configurations, and significantly accelerated inference using GPUs.
We follow the GPTQ 4-bit configuration outlined by \citealp{frantar2022optq} and use a random sample of 128 sequences from the C4 dataset~\cite{raffel2020exploring} for quantization and set a grouping size equal to 128. 
Additional details regarding the quantization procedures can be found in  \autoref{tab:gptq_params}~(\autoref{sec:appendix_1}).

\subsection{Evaluation}
We focus on evaluating models' confidence in predictions before and after quantization in a zero-shot setting.
In an ideal scenario, we expect the model's performance and confidence to remain consistent after quantization, preserving the initial calibration level.
We evaluate the performance of LLMs post-compression using accuracy (Acc.) and calibration error (CE). 
For binary problems, we use the Expected Calibration Error (ECE;\citealp{naeini2015obtaining}), calculated using reliability plots that bin predicted probabilities and compare them against actual accuracy.
In multi-class benchmarks, we use the Adaptive Calibration Error (ACE; \citealp{nixon2019measuring}), which quantifies calibration performance by dividing predictions into equally sized bins based on confidence levels and comparing accuracy and confidence within these subsets.

\begin{table}[h!]
\centering
\footnotesize
\begin{tabular}{l|cccc}
\toprule
{\textbf{Model}} & \textbf{Conf.}  & \textbf{$\text{Conf}_{err}$}  & \textbf{$\text{Conf}_{true}$} & \textbf{H}\\
\midrule
BLOOM & 96.26 & 95.64 & 46.24 & 12.87 \\
+ GPTQ & 96.3 & 95.62 & 45.23$^{*}$ & 12.89 \\
\midrule
OPT & 96.51 & 95.57 & 50.37 & 12.12 \\
+ GPTQ & 96.5 & 95.55 & 49.78$^{*}$ & 12.22 \\
\midrule
Mistral & 96.85 & 95.02 & 61.14 & 10.96 \\
+ GPTQ & 96.89 & 95.13 & 59.73$^{*}$ & 10.87 \\
\midrule
LLaMA &  96.8 & 95.34 & 56.83 & 11.37  \\
+ GPTQ & 96.48 & 95.13 & 53.69$^{*}$ & 12.21$^{*}$ \\
\bottomrule
\end{tabular}
\caption{
Confidence and prediction entropy evaluation results on \textsc{HellaSwag} for LLMs with $\sim$7B parameters.  Quantized LLMs become less confident in both correct and wrong predictions. Conf.:~Mean confidence in predictions;~$\text{Conf}_{err}$:~Mean confidence in wrong predictions;~$\text{Conf}_{true}$:~Mean confidence in true class;~H=Mean predictive entropy in the answers, multiplied by 100. High entropy means that the model is more unsure about its predictions. The $^\star$ denotes a significant difference with a confidence level set at 0.05 (paired $t$-test).}

\label{tab:confidence_score}
\end{table}

Details regarding the binning parameters used are provided in \autoref{sec:appendix_eval}.
We also examine two cases of miscalibration: (1) the model's rejection of correct predictions due to lower confidence and (2) the model's incorrect prediction due to higher confidence.
Specifically, we measure the model's confidence $\text{Conf}_{err}$ when predicting an incorrect class and the model's confidence in the true class $\text{Conf}_{true}$.
\section{Experiment Settings}
\paragraph{Data}
We use six standard commonsense reasoning tasks for our analysis:~\textsc{ARC Easy}~\cite{clark2018think}, \textsc{BoolQ}~\cite{clark2019boolq}, \textsc{PIQA} \cite{bisk2020piqa}, \textsc{HellaSwag}~\cite{zellers2019hellaswag}, \textsc{OBQA}~(OpenBookQA; \citealp{Mihaylov2018CanAS}), and \textsc{XStory-En}~\cite{mostafazadeh2017lsdsem}. 
These benchmarks vary in the types of language inference abilities assessed in LLMs: (1) question answering involving reading comprehension (\textsc{BoolQ}), (2) natural text entailment (\textsc{XStory-En}, \textsc{HellaSwag}), (3) science fact knowledge (\textsc{ARC}, \textsc{OBQA}), and (4) physical commonsense (\textsc{PIQA}).
\paragraph{Models}
We use the following \textit{causal} (auto-regressive) LLMs in our experiments: (1) BLOOM \cite{scao2022bloom}, (2) OPT \cite{zhang2022opt}, (3) Mistral-7B \cite{jiang2023mistral}, and (4) LLaMA-7B \cite{touvron2023llama}. 
To examine how confidence loss varies across different scales, we use various configurations of LLMs: BLOOM with 560M, 1.1B, 1.7B, 3B, and 7.1B parameters, and OPT with 125M, 350M, 1.3B, 2.7B, 6.7B, and 13B parameters. 
 
\section{Results}

\begin{figure}
  \centering
  \includegraphics[width=0.95\linewidth]{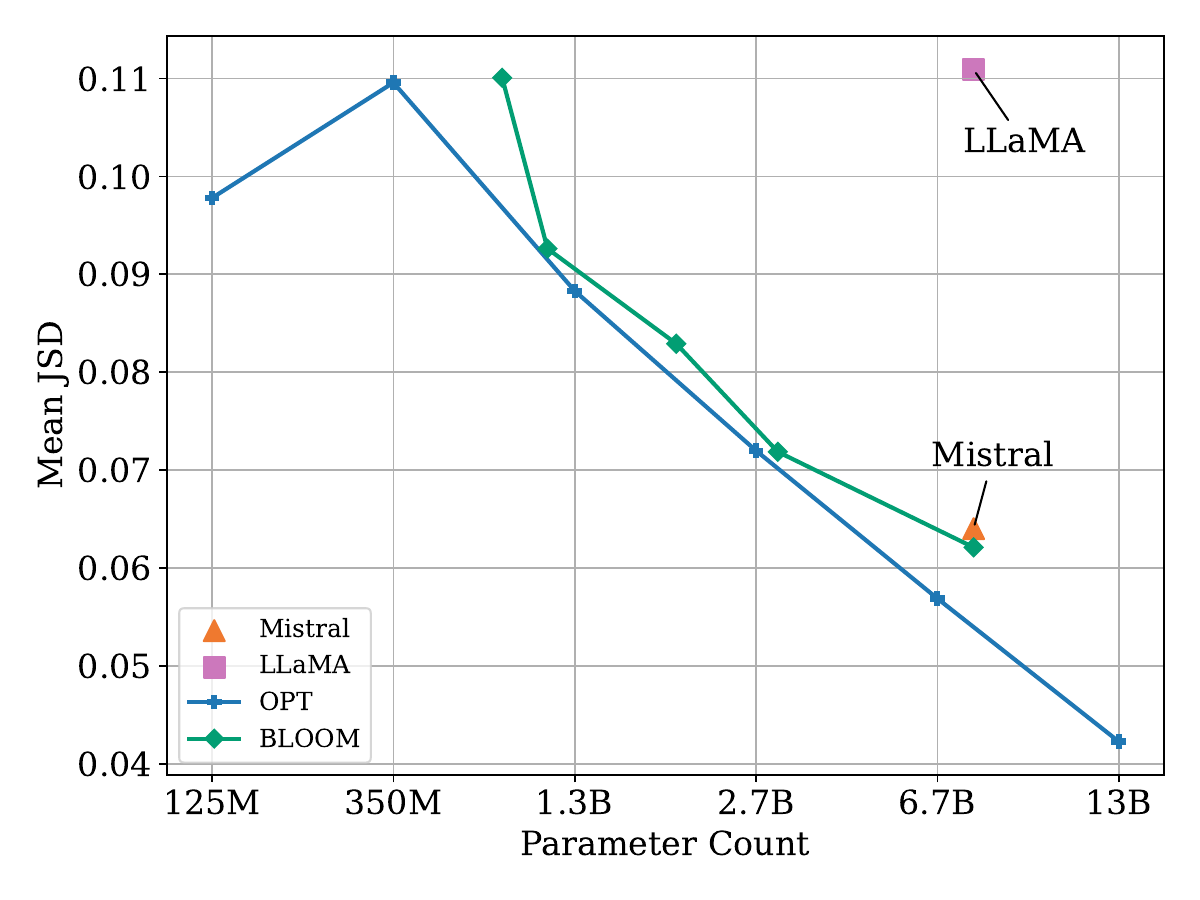}
  \caption{Mean Jensen-Shannon distances between full and quantized LLMs across benchmarks. The distances depict dissimilarities in true-class probability distributions.}
  \label{fig:JSD}
\end{figure}

We conduct a series of experiments to estimate the impact of quantization on various aspects of LLMs' performance, including calibration error, prediction entropy, cases of maximum confidence change, and the distribution dissimilarities between full and compressed models. 
We find variance in quantization impact across different families of models and their sizes, suggesting that scale and pre-training directly affect the further quantization loss.

\paragraph{Calibration Impact}
\autoref{tab:calibration_score} outlines the classification results after quantization, evaluated through calibration error and accuracy metrics, along with the variation of these scores compared to the uncompressed LLMs.
The general trend is that quantization amplifies the pre-existing high calibration error present in the models before compression across different models and benchmarks.
This trend remains consistent across various model families, notably affecting the LLaMA-7B, which experiences a $\sim$10\% increase in pre-compression calibration error on the \textsc{HellaSwag} dataset.
Overall, scores associated with the \textsc{HellaSwag} dataset are more significantly impacted compared to those of the \textsc{BoolQ} and \textsc{PiQA} benchmarks.

\paragraph{Confidence Impact}

\autoref{tab:confidence_score} presents the results obtained from four models, each having a near-equivalent number of parameters. 
Notably, across all models, a consistent trend of overconfidence emerges in both pre- and post-quantization stages, with an average confidence level around $\sim$0.95 for incorrect predictions.
Our analysis further shows a statistically significant impact of quantization on the confidence associated with true-class predictions. 
Additionally, we observe an increase in entropy for the quantized LLMs shown in \autoref{tab:conf_3} (see \autoref{sec:app_d_full_tables}).
This increase suggests an amplification in the variance across answers, reflecting increased uncertainty in answer selection due to quantization.

\paragraph{Identifying Cases of Confidence Change}
To identify instances of confidence change, we segment the models' predictions into bins and calculate the confidence changes after quantization within each bin.
In \autoref{fig:conf_diffs}, we depict the mean confidence changes for the BLOOM and OPT models on the \textsc{HellaSwag} benchmark.
The plot illustrates that samples with lower pre-quantization confidence levels are significantly affected by the quantization process, whereas samples in which the original model was confident show less impact.
This observation suggests that quantization predominantly influences the confidence of samples where the original model exhibited lower confidence levels.

\paragraph{Jensen-Shannon Distances}
To illustrate the extent of differences between the distributions of the full and compressed models, we plot the mean Jensen-Shannon distances across benchmarks in \autoref{fig:JSD}. 
These distances reflect the dissimilarity between the true-class probability distributions of the models.
We find that the distances between original and compressed decrease as the model size scales up. 
Notably, most model families show a consistent trend in this behavior, except for LLaMa, which diverges from the patterns observed in other models of similar size ($\sim$7B).

\section{Conclusion}
This paper investigates the impact of quantization on the confidence and calibration of LLMs. 
We demonstrate that quantization leads to an increase in calibration error and statistically significant changes in confidence levels for correct predictions. 
Through a detailed examination of confidence shifts, we identify instances of confidence change occurring in data where models lack confidence before quantization.
Overall, our findings provide insights into quantization loss and suggest a potential direction for future work, emphasizing the need to focus on calibrating LLMs, specifically on uncertain examples. 
For example, future work may focus on integrating the models' calibration, such as temperature scaling, into the quantization pipeline.
Also, we have demonstrated that different model families, including \textsc{LLaMA}, \textsc{Mistral}, \textsc{BLOOM}, and \textsc{OPT}, exhibit varying degrees of susceptibility to quantization, as measured by changes in confidence levels. 
This suggests another direction for future research – benchmarking LLMs based on their response to quantization-induced confidence shifts.


\section*{Limitations} 
Our quantization techniques are currently limited to 4-bit post-training quantization with GPTQ. 
However, future work can benefit from exploring training-aware quantization approaches, studying different quantization factors, such as 2- and 3-bit weight representation, and quantization of activations.

In our evaluations, we employ zero-shot techniques, enabling the estimation of the pure quantization effect. 
Previous studies mentioned in related work included a fine-tuning step, whereas our approach avoids it. 
Yet, future work could involve few-shot analysis since this method has the potential to amplify or compensate for confidence and quantization loss.

Further research could apply our analysis to other generative tasks. 
Instead of predictive distributions over labels, one could consider those across tokens. 
This means using the full model's predictions as references and comparing the confidence in these generations after the quantization process.

\section*{Acknowledgments}
This work has benefitted from access to the HPC resources provided by IDRIS under the allocation AD011014384, granted by GENCI, which facilitated the utilization of the Jean Zay supercomputer. 
Additionally, this research is supported by the ANR project Diké (No. ANR-21-CE23-0026-02).


\bibliography{anthology,custom}
\clearpage
\newpage
\appendix
\section{Quantization Parameters}\label{sec:appendix_1}
\begin{table}[!h]
    \centering
    \begin{tabular}{lc}
    \toprule
        \textbf{Parameter} & \textbf{Value} \\
        \midrule
        Num bits & 4 \\
        Group size & 128 \\
        Dampening factor (\%) & 0.01 \\
        Desc act & false \\
        Symmetry & true \\
        True sequential & true \\ 
        \bottomrule
    \end{tabular}
    \caption{Configuration for GPTQ}
    \label{tab:gptq_params}
\end{table}

\section{Evaluation Details}\label{sec:appendix_eval}

In this section, we provide further details on the used measures for the experiments.

\paragraph{Jensen-Shannon Divergence}

In Figure~\ref{fig:JSD}, we give the distance dissimilarities in the true-class probability distributions using the Jensen-Shannon divergence.
For a given dataset, we focus on the true class probabilities, $p \in \mathbb{R}^n$, for the full model, and $q \in \mathbb{R}^n$ for the quantized one, where $n$ denotes the number of instances.\\

The Jensen Shannon-Divergence between these two distributions is defined by:

\begin{align*}
        & \; JSD(p,q)  \\ 
    =   & \; \dfrac{1}{2}\left( KL\left( p \mid \mid \dfrac{p + q}{2}\right) + KL\left( q \mid\mid \dfrac{p+ q}{2} \right)\right),\\
    = & \; \sum_{i=1}^n p_i\ln\left(\dfrac{2p_i}{p_i+q_i} \right) + q_i\ln\left(\dfrac{2q_i}{p_i+q_i} \right),
\end{align*}

where $KL$ denoted the Kullback-Leibler divergence and $p_i$ and $q_i$ are the true-class probabilities of the $i$-th instance for the full and quantized model respectively.\\

These distances are then averaged over all the studied datasets.

\paragraph{Expected Calibration Error (ECE)}

Let us consider a model $h$, which assigns confidence (which are probabilities) of belonging in a given class.
These confidence scores can be divided into several bins $B_m, m =1,\ldots M$ where $M$ is the number of bins. 
More precisely, an instance belongs to the bin $B_m$ if its confidence score in the true class $conf_i$ is in a given range (\textit{e.g.} if $(m-1)/M \le conf_i \le m/M$). 
In a given bin $B_m$, we can also measure the accuracy of the model, \textit{i.e.}, compute the ratio of instances in the bin $B_m$ that are well-classified.\\

The \textit{expected calibration error} is then defined as the weighted mean, where the weights depend on the number of instances in the bin of the absolute difference between the accuracy $acc(B_m)$ of the bin and the mean confidence score in the bin $\overline{conf}(B_m) = \dfrac{1}{\vert B_m \vert}\sum_{i \in B_m } conf_i.$, \textit{i.e.},

\[ECE = \sum_{m=1}^M \dfrac{
\vert B_m \vert}{n} \vert acc(B_m) - \overline{conf}(B_m) \vert, \]

where $n$ is the sample size.
Note this error has been developed for binary classification tasks and can be extended to multi-class settings using the so-called \textit{SCE}~\citep{nixon2019measuring}, but this first extension has been shown to be not relevant for all studies~\citep{ulmer-etal-2022-exploring}.
The authors rather use the \textit{adaptive calibration error}, which works with equal size bins.

\paragraph{Adaptive Calibration Error (ACE)}

The \textit{adaptive calibration error} is defined by

\[ACE = \dfrac{1}{CM} \sum_{c=1}^C\sum_{m=1}^M \vert acc(B_m,c) - \overline{conf}(B_m,c) \vert,\]

where $C$ is the number of classes, $M$ is the number of bins that are created, $acc(B_m,c)$ is the accuracy on class $c$ in the $m$-th bin and $\overline{conf}(B_m,c)$ is the mean confidence score for class $c$ in the $m$-th.\\
In this case, all the bins have the same size, which is equal to $\lfloor C/M \rfloor$.

\paragraph{Implementation Details}

Our experiments use evaluation scripts derived from the EleutherAI Language Model Evaluation Harness \cite{eval-harness}.\footnote{\url{https://github.com/EleutherAI/lm-evaluation-harness}}
To quantize the models we use scripts from Auto-GPTQ package.\footnote{\url{https://github.com/PanQiWei/AutoGPTQ}}
We run quantization and inference for all the experiments on a single NVIDIA A-100 GPU.
For the largest model, uncompressed OPT-13B, the evaluation run took roughly two hours for all the datasets.
\citealp{frantar2022gptq} report
GPTQ runtime for the models.
\newpage
\onecolumn 
\section{Confidence Evaluation in LLMs after Quantization} 

In this last experiment we study the evolution of the confidence score for our different models on the six studied datasets.
More precisely, we study the mean difference of confidence score between full and quantized models for different ranges of confidence scores of the full model.\\

\begin{figure*}[!h]
\centering
\includegraphics[width=0.8\textwidth]{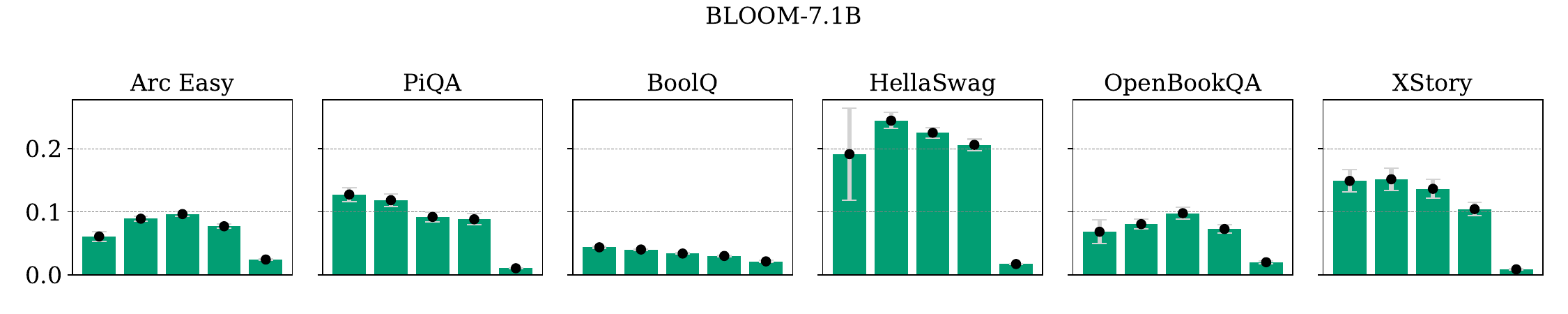}\\
\centering
\includegraphics[width=0.8\textwidth]{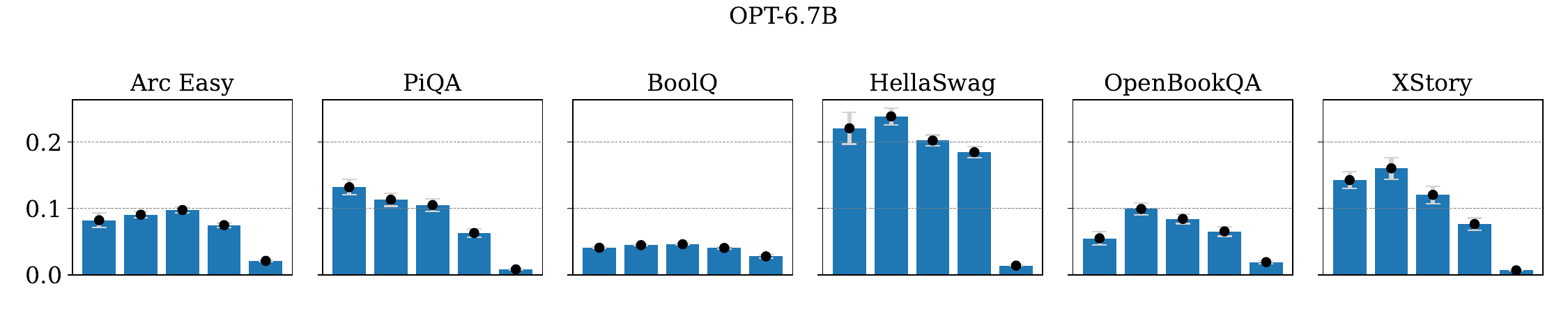}\\
\centering
\includegraphics[width=0.8\textwidth]{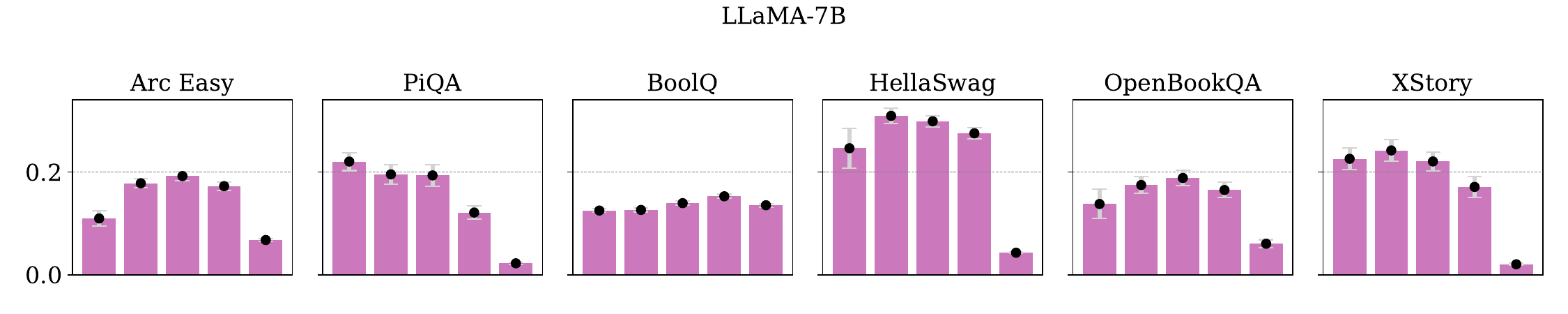}\\
\centering
\includegraphics[width=0.8\textwidth]{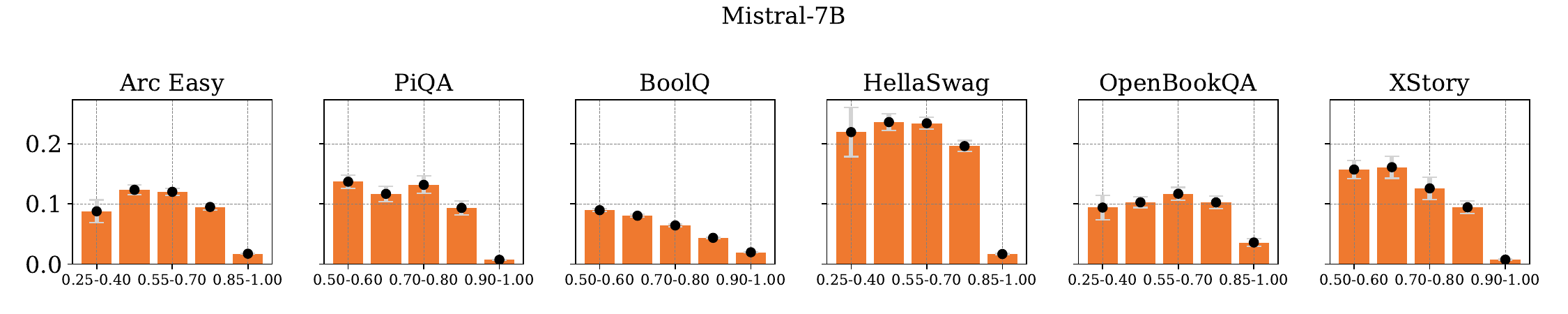}\\
\caption{Confidence difference for models across datasets. 
For each dataset (in column) and each model (in line), we provide the difference in prediction scores between the full and quantized models. 
More precisely, each bar represents the mean difference in confidence between the quantized and full models, with confidence in the full model represented on the horizontal axis.
Note that some ranges start from $0.5$ for binary tasks and $0.25$ for multi-class (with four classes) tasks. 
For a confidence lower than the previous one, there is no chance of being assigned to the associated class.  } 
\label{fig:conf_shift}
\end{figure*}

As presented in \autoref{fig:conf_shift}, the change of probabilities is the lowest one when the model is over-confident and the uncertainty of the model is impacted (\textit{i.e.}, increased) by the quantization. 
This observation goes hand in hand with the entropy values, serving as a measure of model uncertainty, shown in \autoref{tab:conf_3}~(\autoref{sec:app_d_full_tables}).
We also note that, in the case of binary problems (\textsc{Piqa}, \textsc{Boolq} and \textsc{XStory Close En}), that the most impacted confidence scores are the ones for which the model is not confident it its prediction.

\newpage
\section{Confidence Evaluation Results}\label{sec:app_d_full_tables}
\begin{table*}[th!]
{
    \centering
    \footnotesize
    \resizebox{0.95\textwidth}{!}{
        \begin{tabular}{l|cc|cc|cc|cc|cc|cc} 
            \toprule
            \multirow{2}{*}{\textbf{Model}} &
            \multicolumn{2}{c|}{\rotatebox{0}{\textbf{\textsc{Arc Easy}}}} &
            \multicolumn{2}{c|}{\rotatebox{0}{\textbf{\textsc{BoolQ}}}} &
            \multicolumn{2}{c|}{\rotatebox{0}{\textbf{\textsc{HellaSwag}}}} &
            \multicolumn{2}{c|}{\rotatebox{0}{\textbf{\textsc{OpenBookQA}}}} &
            \multicolumn{2}{c|}{\rotatebox{0}{\textbf{\textsc{PiQA}}}} &
            \multicolumn{2}{c}{\rotatebox{0}{\textbf{\textsc{XStory}}}} \\
            \cmidrule{2-13}
            
            & \textbf{Conf.}  & \textbf{$\text{Conf}_{err}$} & \textbf{Conf.}  & \textbf{$\text{Conf}_{err}$} & \textbf{Conf.}  & \textbf{$\text{Conf}_{err}$} & \textbf{Conf.}  & \textbf{$\text{Conf}_{err}$} & \textbf{Conf.}  & \textbf{$\text{Conf}_{err}$} & \textbf{Conf.}  & \textbf{$\text{Conf}_{err}$} \\
            \midrule
            {\includegraphics[height=0.4cm]{images/mistral.png}}{\color{gray}$\cdots$} & \multicolumn{12}{c}{$\cdots$} \\
                        7B & 88.45 & 75.44 & 76.21 & 64.8 & 96.85 & 95.02 & 79.2 & 78.93 & 94.78 & 89.39 & 95.23 & 95.57\\
            7BQ & 88.15 & 75.61 & 76.75$^*$ & 65.56 & 96.89 & 95.13 & 79.01 & 79.07 & 94.67 & 88.8 & 95.26 & 95.72\\
            {\includegraphics[height=0.5cm]{images/llama.png}}{\color{gray}$\cdots$} & \multicolumn{12}{c}{$\cdots$}\\
                        7B & 84.68 & 73.12 & 75.75 & 67.78 & 96.8 & 95.34 & 78.64 & 78.85 & 94.24 & 90.32 & 95.01 & 95.51 \\
            7BQ & 81.6$^*$ & 71.93 & 68.95$^*$ & 63.55 & 96.48$^*$ & 95.13 & 78.25 & 78.68 & 93.99 & 90.05 & 94.8 & 94.94\\            {\includegraphics[height=0.5cm]{images/cherry_blossom.png}}{\color{gray}$\cdots$} & \multicolumn{12}{c}{$\cdots$} \\
            560M & 76.45 & 73.76 & 64.74 & 64.38 & 96.39 & 96.27 & 78.28 & 78.77 & 91.47 & 89.8 & 94.47 & 94.62\\
            560MQ & 75.89 & 73.68 & 61.89$^*$ & 62.76 & 96.47 & 96.35 & 78.74 & 79.11 & 91.76 & 90.07 & 94.55 & 95.22 \\
            1.1B & 76.2 & 72.22 & 70.63 & 69.16 & 96.45 & 96.16 & 78.3 & 78.81 & 92.1 & 89.99 & 94.36 & 94.46 \\
            1.1BQ & 76.0 & 72.95 & 73.28$^*$ & 72.08 & 96.52 & 96.18 & 77.97 & 78.31 & 91.78 & 89.58 & 94.21 & 94.87 \\
            1.7B & 77.47 & 72.86 & 76.12 & 74.9 & 96.24 & 95.89 & 78.05 & 78.57 & 91.89 & 89.75 & 93.96 & 94.08\\
            1.7BQ  & 76.34$^*$ & 72.44 & 76.06 & 74.85 & 96.11 & 95.91 & 77.52 & 78.31 & 91.54 & 89.56 & 93.87 & 94.32\\
            3B & 78.6 & 73.11& 72.5 & 70.55 & 96.24 & 95.75 & 78.3 & 78.56 & 92.37 & 89.57 & 94.36 & 94.08 \\
            3BQ & 77.24$^*$ & 72.23& 71.59$^*$ & 69.93& 96.43 & 96.02& 77.5 & 77.9& 92.25 & 89.2& 94.15 & 94.32 \\
            7.1B & 79.95 & 73.11& 69.97 & 66.55& 96.26 & 95.64 & 78.36 & 78.52 & 92.71 & 88.05 & 94.59 & 94.85 \\
            7.1BQ & 79.46 & 72.85 & 69.59$^*$ & 66.58 & 96.3 & 95.62 & 78.17 & 78.59& 92.53 & 89.03 & 94.56 & 94.54 \\
            {\includegraphics[height=0.5cm]{images/opt3.png}}{\color{gray}$\cdots$} & \multicolumn{12}{c}{$\cdots$} \\ 
            125M & 75.9 & 74.29 & 67.95$^*$ & 67.42 & 96.31 & 96.29 & 77.6 & 78.42 & 90.96 & 89.31 & 94.1 & 94.86\\
            125MQ  & 75.88 & 74.57 & 64.45 & 64.21 & 96.29 & 96.15 & 78.14 & 79.62 & 91.39 & 89.89 & 94.31 & 94.59\\
            350M & 75.45 & 73.25 & 67.45 & 66.57 & 96.07 & 95.91 & 78.36 & 79.38 & 91.39 & 88.85 & 94.17 & 94.62\\
            350MQ & 76.46$^*$ & 74.84 & 63.03$^*$ & 62.34 & 96.25 & 96.03 & 78.26 & 78.68 & 91.33 & 89.23 & 94.38 & 94.43\\
            1.3B & 77.67 & 72.44 & 64.25 & 62.24 & 96.31 & 95.74 & 78.35 & 79.07 & 91.91 & 88.69 & 94.39 & 94.47  \\
            1.3BQ  & 77.02$^*$ & 72.54 & 64.26 & 63.5 & 96.14 & 95.58 & 78.56 & 79.16 & 91.94 & 88.59 & 94.31 & 94.87 \\
            2.7B & 78.22 & 71.73 & 63.67 & 61.81 & 96.32 & 95.51 & 78.45 & 78.68 & 91.89 & 87.93 & 94.42 & 94.6\\
            2.7BQ & 77.58$^*$ & 71.69 & 63.66 & 62.07 & 96.2 & 95.62 & 77.89 & 77.66 & 92.14 & 88.02 & 94.35 & 94.64\\
            6.7B & 80.46 & 72.14 & 65.88 & 62.46 & 96.51 & 95.57 & 78.65 & 79.16 & 93.29 & 89.78 & 94.38 & 95.32\\
            6.7BQ & 80.29 & 72.52 & 64.16$^*$ & 60.9 & 96.5 & 95.55 & 78.66 & 78.32 & 93.13 & 89.4 & 94.55 & 94.69\\
            13B & 81.36 & 72.42 & 67.3 & 63.32 & 96.49 & 95.48 & 78.7 & 78.75 & 93.23 & 88.64 & 94.98 & 95.53\\
            13BQ & 80.96$^*$ & 72.4 & 66.78$^*$ & 62.35 & 96.5 & 95.52 & 79.08 & 79.46 & 93.03 & 88.77 & 94.77 & 95.18\\
            \bottomrule
        \end{tabular}}}
    \caption{\label{tab:conf_2} 
    Mean confidence evaluation results across benchmarks. Conf.:~Mean confidence in predictions;~$\text{Conf}_{err}$:~Mean confidence in wrong predictions. The $^\star$ is used to denote a significant difference with a confidence level set at 0.05 (paired $t$-test). Q denotes quantized models. \underline{Notations:}
    \includegraphics[height=0.35cm]{images/mistral.png}=\textsc{Mistral}; \includegraphics[height=0.35cm]{images/llama.png}=\textsc{LLaMA};\includegraphics[height=0.35cm]{images/cherry_blossom.png}=\textsc{BLOOM};\includegraphics[height=0.35cm]{images/opt3.png}=\textsc{OPT}.}
\end{table*}

\begin{table*}[th!]
{
    \centering
    \footnotesize
    \resizebox{0.95\textwidth}{!}{
        \begin{tabular}{l|cc|cc|cc|cc|cc|cc} 
            \toprule
            \multirow{2}{*}{\textbf{Model}} &
            \multicolumn{2}{c|}{\rotatebox{0}{\textbf{\textsc{Arc Easy}}}} &
            \multicolumn{2}{c|}{\rotatebox{0}{\textbf{\textsc{BoolQ}}}} &
            \multicolumn{2}{c|}{\rotatebox{0}{\textbf{\textsc{HellaSwag}}}} &
            \multicolumn{2}{c|}{\rotatebox{0}{\textbf{\textsc{OpenBookQA}}}} &
            \multicolumn{2}{c|}{\rotatebox{0}{\textbf{\textsc{PiQA}}}} &
            \multicolumn{2}{c}{\rotatebox{0}{\textbf{\textsc{XStory}}}} \\
            \cmidrule{2-13}
            & \textbf{$\text{Conf}_{true}$} & \textbf{H} & \textbf{$\text{Conf}_{true}$} & \textbf{H} & \textbf{$\text{Conf}_{true}$} & \textbf{H} & \textbf{$\text{Conf}_{true}$} & \textbf{H} & \textbf{$\text{Conf}_{true}$} & \textbf{H} & \textbf{$\text{Conf}_{true}$} & \textbf{H} \\
            
            \midrule
            {\includegraphics[height=0.4cm]{images/mistral.png}}{\color{gray}$\cdots$} & \multicolumn{12}{c}{$\cdots$} \\
            7B &  76.82 & 43.09 & 71.37 & 71.05 & 61.14 & 10.96 & 30.93 & 72.41 & 79.52 & 17.69 & 47.4 & 16.05\\
            7BQ & 75.76$^*$ & 44.42$^*$ & 71.37 & 70.07$^*$ & 59.73$^*$ & 10.87 & 30.36 & 73.19 & 79.14 & 18.11 & 47.27 & 16.1\\{\includegraphics[height=0.5cm]{images/llama.png}}{\color{gray}$\cdots$} & \multicolumn{12}{c}{$\cdots$}\\
            7B & 70.22 & 56.26 & 66.83 & 71.19 & 56.83 & 11.37 & 31.28 & 73.87 & 77.04 & 19.53 & 46.88 & 16.62 \\
            7BQ & 65.01$^*$ & 66.4$^*$ & 61.48$^*$ & 84.3$^*$ & 53.69$^*$ & 12.21$^*$ & 28.42$^*$ & 76.09 & 75.3$^*$ & 20.61$^*$ & 46.97 & 17.65$^*$ \\{\includegraphics[height=0.5cm]{images/cherry_blossom.png}}{\color{gray}$\cdots$} & \multicolumn{12}{c}{$\cdots$} \\
            560M  & 44.01 & 82.38 & 51.83 & 90.77 & 31.5 & 12.41 & 17.68 & 75.57 & 62.89 & 27.68 & 47.93 & 18.72\\
            560MQ & 42.42$^*$ & 83.81$^*$ & 49.78$^*$ & 93.37$^*$ & 31.07$^*$ & 12.26 & 17.65 & 73.97 & 62.07 & 27.52 & 48.16 & 18.61 \\
            1.1B & 47.33 & 83.0 & 54.95 & 82.92 & 34.51 & 12.35 & 19.62 & 75.19 & 65.82 & 25.78 & 48.11 & 19.33\\
            1.1BQ  & 45.23$^*$ & 83.28 & 55.5$^*$ & 79.2$^*$ & 33.67$^*$ & 12.28 & 18.26$^*$ & 76.23 & 64.99$^*$ & 26.44 & 48.3 & 19.66 \\
            1.7B & 51.15 & 79.99 & 57.09 & 74.65 & 37.52 & 13.11 & 21.19 & 75.97 & 67.06 & 26.28 & 46.93 & 19.99\\
            1.7BQ  & 49.21$^*$ & 83.58$^*$ & 56.95 & 74.95 & 36.52$^*$ & 13.38 & 19.83$^*$ & 77.84$^*$  & 66.23$^*$ & 27.16$^*$  & 47.41 & 20.4\\
            3B & 54.29 & 76.14 & 56.73 & 79.88 & 41.26 & 12.91 & 22.3 & 74.9 & 69.3 & 25.11 & 47.1 & 19.1 \\
            3BQ & 52.07$^*$ & 80.42$^*$ & 56.34$^*$ & 81.98$^*$ & 40.41$^*$ & 12.55$^*$ & 21.64 & 77.03$^*$ & 68.75 & 25.6 & 47.57 & 19.55\\
            7.1B & 59.1 & 71.97 & 57.67 & 83.21 & 46.24 & 12.87 & 24.64 & 75.42 & 71.88 & 23.84 & 46.8 & 18.6 \\
            7.1BQ & 57.79$^*$ & 73.52$^*$ & 57.33$^*$ & 83.89$^*$ & 45.23$^*$ & 12.89 & 23.74$^*$ & 75.9 & 71.38 & 24.41$^*$  & 46.69 & 18.65 \\
            {\includegraphics[height=0.5cm]{images/opt3.png}}{\color{gray}$\cdots$} & \multicolumn{12}{c}{$\cdots$} \\ 
            125M & 40.09 & 83.61 & 52.46 & 86.7 & 29.04 & 12.86 & 16.95 & 77.19 & 61.88 & 28.97$^*$  & 48.26 & 20.12\\
            125MQ & 39.12$^*$ & 83.58 & 50.99$^*$ & 91.0$^*$ & 28.62$^*$ & 12.75 & 16.27 & 76.26 & 61.22 & 28.24 & 48.33 & 19.54\\
            350M & 41.01 & 85.36 & 53.4 & 87.14 & 32.03 & 13.51 & 17.41 & 76.08 & 63.78 & 28.0 & 47.69 & 19.77\\
            350MQ & 40.69 & 82.59$^*$ & 51.72$^*$ & 92.36$^*$ & 31.84 & 13.09 & 17.05 & 75.09 & 62.73$^*$ & 28.31 & 48.04 & 19.16\\
            1.3B & 52.33 & 79.37 & 53.88 & 90.87 & 41.5 & 13.03 & 22.51 & 75.63 & 70.02 & 26.08 & 47.12 & 18.73\\
            1.3BQ  & 50.83$^*$ & 81.23$^*$ & 52.07$^*$ & 91.22 & 40.47$^*$ & 13.26 & 22.2 & 75.09 & 69.52 & 25.96 & 47.21 & 18.76 \\
            2.7B & 55.63 & 77.64 & 54.34 & 91.63 & 45.84 & 12.84 & 24.63 & 74.97 & 72.0 & 25.68 & 46.89 & 18.82\\
            2.7BQ & 53.91$^*$ & 79.47$^*$ & 52.88$^*$ & 91.66 & 45.12$^*$ & 13.09 & 23.71 & 76.5 & 71.45 & 25.45 & 46.9 & 18.78\\
            6.7B & 60.58 & 70.14 & 57.42 & 88.65 & 50.37 & 12.12 & 26.42 & 74.26 & 74.42 & 22.67 & 46.81 & 18.3\\
            6.7BQ  & 60.08 & 70.89 & 56.57$^*$ & 90.78$^*$ & 49.78$^*$ & 12.22 & 26.45 & 74.65 & 74.26 & 22.96 & 46.63 & 18.06\\
            13B & 62.35 & 67.09 & 58.24 & 86.84 & 52.15 & 12.25 & 26.85 & 73.83 & 74.56 & 22.26 & 46.88 & 17.13\\
            13BQ & 62.06 & 68.26$^*$ & 58.38 & 87.45$^*$ & 51.61$^*$ & 12.24 & 26.67 & 73.7 & 74.38 & 22.69$^*$  & 46.98 & 17.38\\
            \bottomrule
        \end{tabular}}}
    \caption{\label{tab:conf_3} 
    Mean confidence in true classes and predictive entropy evaluation results across benchmarks. ~$\text{Conf}_{true}$:~Mean confidence in true class;~H=Mean Predictive entropy in the answers, multiplied by 100. The $^\star$ is used to denote a significant difference with a confidence level set at 0.05 (paired $t$-test).
    Q denotes quantized models.
    \underline{Notations:}
    \includegraphics[height=0.35cm]{images/mistral.png}=\textsc{Mistral}; \includegraphics[height=0.35cm]{images/llama.png}=\textsc{LLaMA};\includegraphics[height=0.35cm]{images/cherry_blossom.png}=\textsc{BLOOM};\includegraphics[height=0.35cm]{images/opt3.png}=\textsc{OPT}.}
\end{table*}
\end{document}